\documentclass[letterpaper, 10 pt, conference]{ieeeconf}
\IEEEoverridecommandlockouts
\usepackage{amsmath}
\usepackage{booktabs}
\usepackage{graphicx}
\usepackage{cite}
\usepackage{url}
\usepackage{xspace}

\newcommand{\NavPatch}{\textsc{NavPatch}\xspace}
\newcommand{\ADD}{\textsc{Add}\xspace}
\newcommand{\REMOVE}{\textsc{Remove}\xspace}
\newcommand{\EXTEND}{\textsc{Extend}\xspace}

\title{\NavPatch: Evidence-Guided Object-Level Costmap Correction with Vision–Language Models}
\author{Shiji Sun$^{1}$, Xingyu Tao$^{2}$, Hao Wang$^{1}$, Ling Wang$^{3,*}$, and Zhengyi Chen$^{1,*}$%
\thanks{$^{1}$Department of Intelligent Construction and Operation, Southeast University, Nanjing 211189, China.}%
\thanks{$^{2}$Department of Construction Management and Intelligence, The Hong Kong Polytechnic University, Hong Kong SAR, China.}%
\thanks{$^{3}$Department of Automation, Tsinghua University, Beijing 100084, China.}%
\thanks{$^{*}$Corresponding authors: Ling Wang (wangling@tsinghua.edu.cn) and Zhengyi Chen (zhengyi.chen@seu.edu.cn).}%
}

\begin{document}
\maketitle
\thispagestyle{empty}
\pagestyle{empty}

\begin{abstract}
Mobile robots typically rely on geometric maps for obstacle avoidance and path planning, but the resulting obstacle representation does not always match how an object should affect navigation. A low lying cable may be missed, a flexible curtain may create spurious blockage, and a traffic cone may require an exclusion region larger than its observed footprint. We present NavPatch, an object level correction layer that assigns ADD, REMOVE, or EXTEND to navigation relevant object categories through periodic scene understanding with a vision--language model. Open vocabulary grounding localizes object instances, and LiDAR and RGB-D observations provide 3D support. Observation quality filtering and cross frame maintenance determine when each correction patch is committed, replaced, or revoked. In 50 real robot trials across five layouts, NavPatch achieves an overall success rate of $86.0\%$. An ablation study of four configurations with 200 runs in total shows that NavPatch improves the success rate from $70.0\%$ to $86.0\%$ and reduces the false commit rate from $68.4\%$ to $40.7\%$ compared with updates based only on the current observation.
\end{abstract}

\section{Introduction}
\begin{figure}[!t]
  \centering
  \includegraphics[width=\columnwidth]{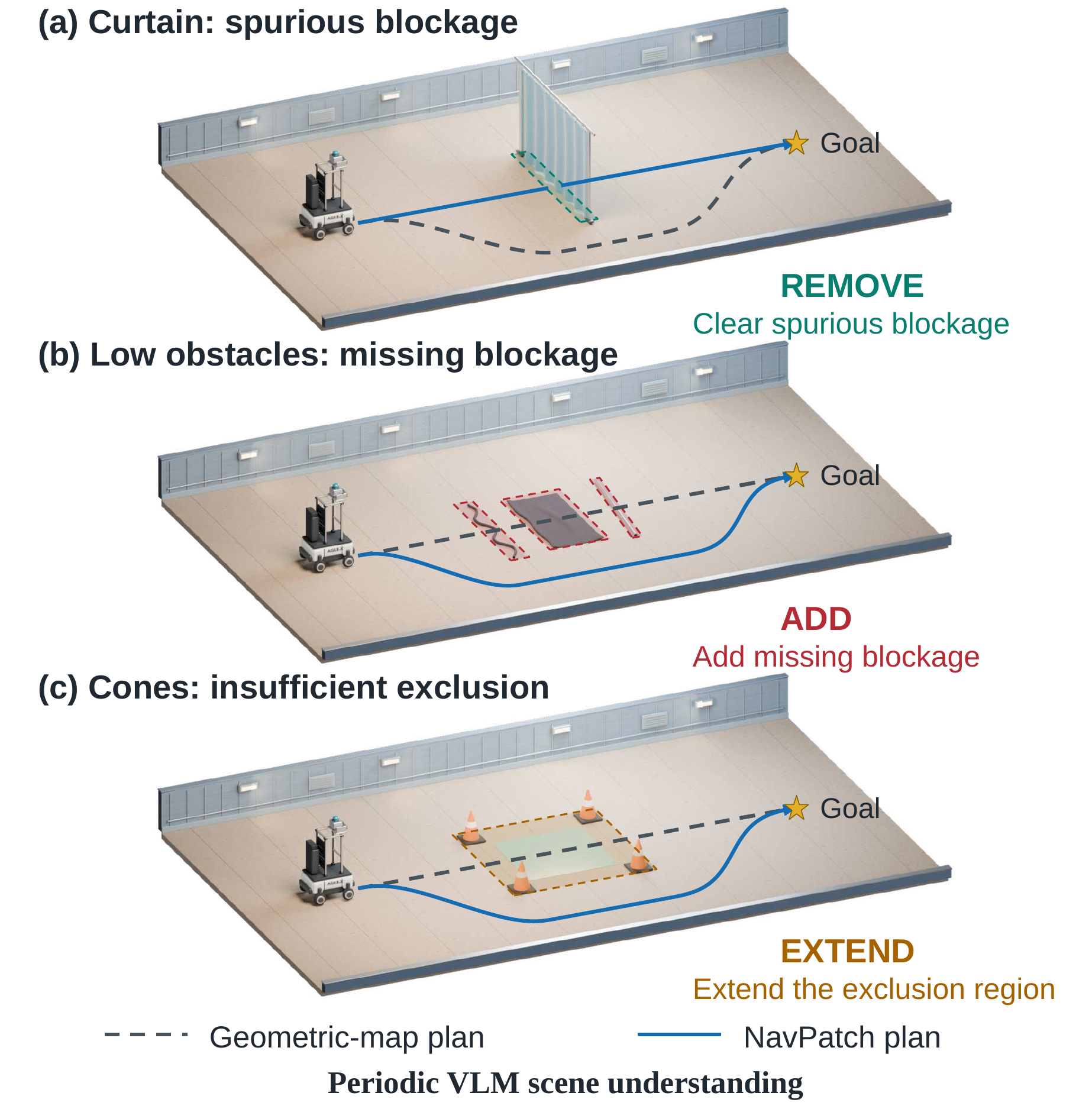}
  \caption{Three representative geometry and navigation mismatches. (a) A traversable curtain creates spurious blockage, (b) a low profile obstacle is weakly represented or missed, and (c) traffic cones require an exclusion region beyond their physical footprint. \NavPatch applies \REMOVE, \ADD, and \EXTEND, respectively. Dashed and solid paths show the plans before and after map correction.} 
  \label{fig:teaser}
\end{figure}

Geometric costmaps remain the primary representation used by many mobile robot planners because they provide a simple and reliable interface for collision avoidance and path planning~\cite{lu2014layered}. However, geometric occupancy describes what is physically observed rather than how an object should affect robot motion. When these two meanings disagree, the costmap can provide an incorrect local constraint to the planner. As illustrated in Fig.~\ref{fig:teaser}, a low profile obstacle may be weakly represented or missed, a traversable curtain may appear as a blocking obstacle, and traffic cones may require a keep out region larger than their physical footprint. These examples motivate the three types of geometry and navigation mismatches considered in this work, missing blockage, spurious blockage, and insufficient exclusion extent.

Recent methods derive navigation constraints from different sources. Interactive Navigation infers object traversability from language and vision~\cite{zhang2024interactive}, while BehAV and NORM-Nav convert language-based behavioral rules into navigation costs~\cite{weera2025behav,huo2026normnav}. E2Map updates navigation costs after experienced events~\cite{kim2025e2map}. Language as Cost instead analyzes the current visual scene to identify potential hazards and update risk costs during navigation~\cite{oh2025languagecost}. We study how scene understanding can support different local map corrections within the same navigation task. The required correction may add missing blockage, clear a permitted passage, or extend an exclusion region. The challenge is to localize these corrections to individual objects and determine when noisy observations justify updating the costmap.

We propose \NavPatch, an object level correction framework for geometric costmaps. A vision--language model (VLM) periodically interprets the full RGB scene throughout navigation to identify relevant object categories and select their map actions. \ADD supplements missing obstacle regions, \REMOVE clears spurious blockage, and \EXTEND enlarges exclusion regions beyond the observed object footprint. Open vocabulary grounding localizes individual objects, while LiDAR and RGB-D observations provide the 3D support used to construct local correction regions. Because the observed object extent can change with viewpoint and localization can fail, each correction is associated with an object instance and supported by evidence across frames. Confirmed regions are maintained as correction patches that can be retained, replaced, or revoked. The resulting patches are fused with the geometric costmap for planning.

The main contributions of this paper are as follows:
\begin{itemize}
  \item We introduce an object-level costmap correction framework that translates periodic VLM scene understanding into ADD, REMOVE, and EXTEND operations to address missing blockage, spurious blockage, and insufficient exclusion extent.
  \item  We develop quality-aware geometric grounding that converts category-level proposals into instance specific correction regions through action conditioned spatial construction and modality specific quality filtering of LiDAR and RGB-D observations. 
  \item  We design an evidence guided patch maintenance mechanism that accumulates quality weighted observations across frames to control the commitment and replacement of object-level costmap corrections.
\end{itemize}

\section{Related Work}

\subsection{VLM for Robot Navigation}
Large VLMs and open vocabulary vision models have been widely adopted in robot navigation. VLFM and OpenFMNav use vision language features or foundation model semantics to estimate where a target is more likely to appear and guide exploration in unknown environments~\cite{yokoyama2024vlfm,kuang2024openfmnav}. ReasonNav organizes doors, signs, people, and map frontiers as navigation landmarks and uses a VLM for high level behavioral reasoning~\cite{chandaka2025reasonnav}. Another class of methods uses VLMs more directly for motion decisions. NaVILA generates mid-level navigation actions from visual observations and language instructions~\cite{cheng2025navila}, whereas CoNVOI selects scene appropriate local trajectories from geometrically feasible candidate positions~\cite{sathyamoorthy2024convoi}. In these methods, semantic reasoning primarily supports object search, behavior selection, or motion decision making.

VLM reasoning has also been encoded as map constraints that can be used directly by planners. Interactive Navigation infers object traversability from language and vision, and combines open vocabulary detection with LiDAR to construct an action aware costmap~\cite{zhang2024interactive}. BehAV maps behavioral rules expressed in natural language to spatial costs~\cite{weera2025behav}, while NORM-Nav uses multilayer costmaps to represent object semantics, direction, speed, and traversability constraints~\cite{huo2026normnav}. These methods use traversability specified through language or behavioral constraints to determine how particular objects should affect navigation. VLM-GroNav focuses on terrain traversability and updates its predictions using physical information acquired during robot motion~\cite{elnoor2025vlmgronav}. E2Map updates navigation costs in response to experienced events~\cite{kim2025e2map}. Language as Cost repeatedly analyzes visual scenes for potential hazards and fuses the resulting risk costs with an obstacle map while retaining geometric blockage~\cite{oh2025languagecost}. \NavPatch uses scene understanding during navigation, but applies distinct local map operations. \ADD supplements missing obstacles, \REMOVE clears blockage caused by permitted traversable objects from the online obstacle layer, and \EXTEND enlarges exclusion regions. These corrections are maintained as patches associated with individual objects.

\subsection{Semantic and Traversability Maps}
Semantic maps combine open vocabulary semantics from visual models with spatial robot representations so that objects can be localized and persistently queried. VLMaps fuses vision language features into a 3D map for open vocabulary landmark queries and language guided navigation~\cite{huang2023vlmaps}. OpenScene and ConceptFusion align open vocabulary or multimodal features with 3D space~\cite{peng2023openscene,jatavallabhula2023conceptfusion}, while ConceptGraphs and HOV-SG further organize open vocabulary scene representations around objects and their spatial relationships~\cite{gu2024conceptgraphs,werby2024hovsg}. OneMap maintains a reusable open vocabulary semantic map for continuous search for multiple goals~\cite{busch2025onemap}, and Hydra demonstrates the feasibility of online 3D scene graphs for persistent spatial perception~\cite{hughes2022hydra}.

As robots move from one shot mapping toward persistent operation, recent work has also considered association and update of the same object across observations acquired at different times. DualMap continuously updates open vocabulary objects through global and local maps in dynamic environments~\cite{jiang2025dualmap}, and SuperMap combines asynchronous open vocabulary perception with geometric SLAM to maintain object identities and semantic states over time~\cite{zhao2026supermap}. ApexNAV preserves target relevant semantic memory through target centric semantic fusion~\cite{zhang2025apexnav}. For traversability representation, WVN estimates visual traversability through online self supervised learning~\cite{frey2023wvn}, and VERN fuses vegetation categories with LiDAR into a vegetation aware costmap~\cite{sathyamoorthy2023vern}. In \NavPatch, object association serves to maintain the correction applied to the navigation map. Observations of the same instance determine whether its patch should be committed, replaced, or revoked.

\section{Problem Formulation}
We consider navigation of a mobile robot in static environments. The robot uses localization, a static map, and online geometric obstacle updates to maintain a base geometric costmap $C_t^{\mathrm{geo}}$. We focus on local regions where the navigation role of a visually perceived object is inconsistent with this geometric representation and correct these regions. We assume that objects requiring correction remain approximately stationary during a single navigation episode.

For a navigation relevant object category in the current scene, the VLM outputs a correction proposal
\begin{equation}
  y_k^t=(c_k^t,p_k^t,a_k^t,s_k^t),
  \label{eq:proposal}
\end{equation}
where $c_k^t$ denotes the object category, $p_k^t$ the detection phrase used for open vocabulary instance localization, $a_k^t$ the map action, and $s_k^t$ the VLM confidence. The \EXTEND action additionally includes direction and extent. 

Each valid object observation yields a local correction region $R_i^t$, represented as a set of 2D grid cells in the map frame. A region denotes an observation level correction, whereas a patch denotes a confirmed region committed to the map. The active correction patches form the set $P_t$, which is applied to the base geometric navigation map:
\begin{equation}
  C_t^{\mathrm{nav}}
  =\mathcal{U}\!\left(C_t^{\mathrm{geo}},P_t\right),
  \label{eq:map_update}
\end{equation}
where $C_t^{\mathrm{nav}}$ is the corrected navigation map used by the downstream planner, $C_t^{\mathrm{geo}}$ is the base costmap composed of the static map and online LiDAR obstacles.

\begin{figure*}[t]
  \centering
  \includegraphics[width=\textwidth]{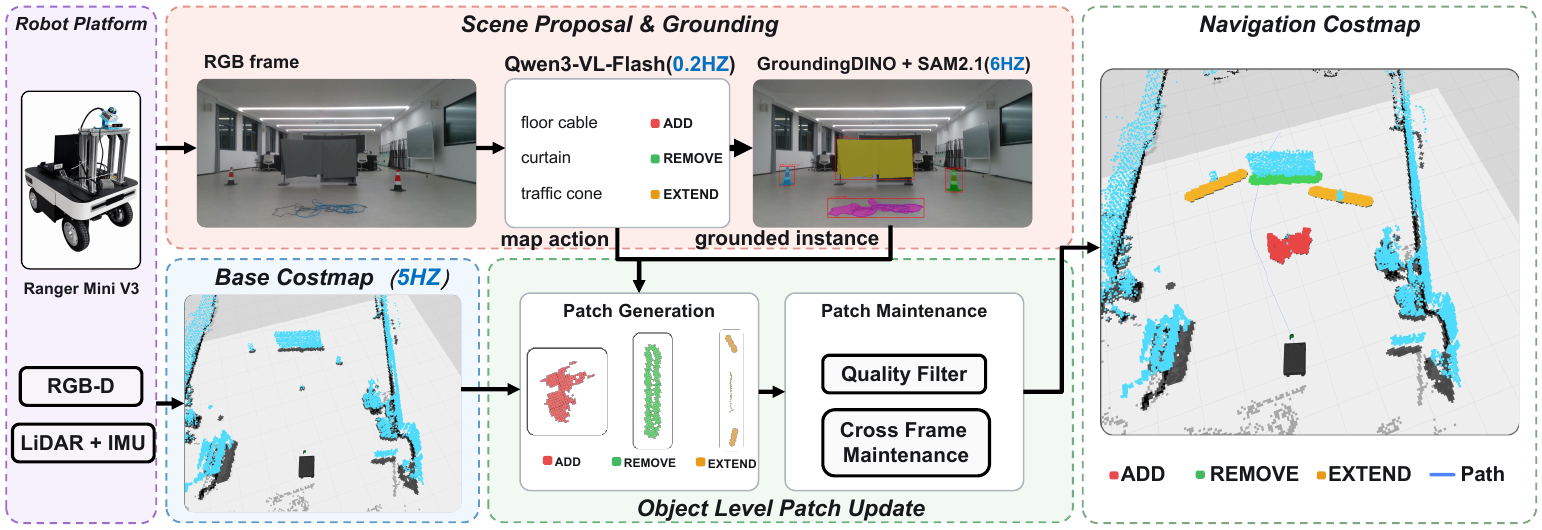}
  \caption{System overview of \NavPatch. Periodic VLM scene understanding proposes object categories and map actions. GroundingDINO and SAM~2.1 localize instances, while LiDAR and RGB-D observations provide 3D support for local correction regions. Observation quality filtering and cross frame maintenance regulate patch updates before correction patches are fused with the base costmap for navigation.}
  \label{fig:framework}
\end{figure*}

\section{Method}

\subsection{System Overview}
The architecture of \NavPatch is shown in Fig.~\ref{fig:framework}. FAST-LIO2~\cite{xu2022fastlio2} takes LiDAR and IMU measurements to provide continuous pose estimates. The navigation system uses a layered costmap, in which the static map and STVL~\cite{macenski2020stvl} provide static and  geometric obstacles, respectively. In parallel, the VLM periodically interprets the full RGB scene during navigation to identify navigation relevant object categories and their map actions. GroundingDINO~\cite{liu2024groundingdino} and SAM~2.1~\cite{ravi2025sam2} localize object instances in subsequent images, while LiDAR and aligned depth provide 3D support for each instance. Current instance observations first undergo instance association and observation quality filtering. Accepted observations are then passed to cross frame maintenance of instance patches. The resulting correction patches are finally fused into the navigation costmap, and the planner directly operates on the updated costmap.

\subsection{Periodic VLM Scene Understanding}
The VLM branch of \NavPatch remains active throughout navigation and periodically processes the current full RGB scene using Qwen3-vl-flash at $0.2\,\mathrm{Hz}$. A structured prompt asks the model to identify navigation relevant object categories in the current scene and to output a detection phrase, map action, and confidence for each category. For \EXTEND, the VLM also outputs a direction and a predefined extent level. 
A fixed confidence threshold of $0.65$ is used throughout all experiments to reject low confidence proposals. Only one current correction proposal is retained for each object category, and it remains active until the next VLM update.

\subsection{Open Vocabulary Instance Localization and 3D Point Extraction}
For each active correction proposal, \NavPatch uses GroundingDINO to localize the corresponding objects in subsequent RGB images according to the detection phrase $p_k^t$, and uses the resulting bounding boxes as prompts to SAM~2.1 to obtain instance masks. GroundingDINO and SAM~2.1 run at approximately $6\,\mathrm{Hz}$ during the real robot experiments. Multiple detections of the same object category are treated as independent object instances. LiDAR, aligned depth, and the corresponding pose transformation are synchronized to the timestamp of the detection image, and the instance mask is used to extract 3D observations within the object region. For LiDAR, projected points falling inside the mask are taken as candidates. For aligned depth, depth pixels within the valid range and covered by the mask are projected back into 3D. The resulting points are grouped by 3D Euclidean clustering with a $0.12\,\mathrm{m}$ distance threshold, and clusters with fewer than three points are discarded. Among the remaining clusters, the one with the smallest median distance to the camera is selected. If it contains at least eight points, the selected cluster is retained as the current 3D support for that modality.

\subsection{Action Conditioned Local Map Correction}
Each localized instance inherits the map action from its corresponding category level VLM proposal. For each available modality $m$, \NavPatch applies the action specific spatial construction to its 3D support and rasterizes the result into a local correction region $R_{i,m}^t$, represented as a set of 2D grid cells in the map frame.

\ADD targets physical obstacles that should block the robot but are weakly or incompletely represented in the geometric map. The currently observed LiDAR and aligned depth 3D points are separately projected onto the ground plane and rasterized into candidate obstacle regions.

\REMOVE targets compliant objects that geometric sensing treats as obstacles but the navigation rule permits the robot to traverse. Only LiDAR points associated with the instance mask are used. The 3D points are rasterized, and small gaps in the grid region are filled by dilation followed by erosion. A fixed margin is then added around the region to form the candidate clearing region.

\EXTEND is used when an object requires a keep out region larger than its physical footprint. The VLM selects one of three predefined modes. AROUND creates a disc around the object, ACROSS creates a strip centered on the object and perpendicular to the line connecting the robot and the object, and BEHIND creates a rectangle extending from the object away from the robot. It also selects a small, medium, or large extent, corresponding to $0.30$, $0.80$, or $1.50\,\mathrm{m}$. Using an anchor derived from the observed 3D points, \NavPatch constructs the region according to the selected mode and extent and rasterizes it into a candidate exclusion region.

\subsection{Observation Quality Filtering}
The 3D support of an object instance varies with viewpoint, image boundary truncation, and point cloud distribution~\cite{yokoyama2024vlfm,zhang2025apexnav}. \NavPatch evaluates LiDAR and aligned depth separately and uses the resulting quality score to determine whether each modality specific correction region is retained for the current observation. For instance $i$ at time $t$, the quality score for modality $m$ is
\begin{equation}
q_{i,m}^{t}
=
q_{\mathrm{bnd},i}^{t}
q_{\mathrm{pts},i,m}^{t}
q_{\mathrm{clu},i,m}^{t},
\label{eq:q_total}
\end{equation}
where $q_{\mathrm{bnd}}$ reflects the fraction of mask pixels outside the image border band, $q_{\mathrm{pts}}$ assesses whether sufficient 3D point support is available, and $q_{\mathrm{clu}}$ measures how strongly the mask associated 3D points concentrate in the selected cluster. A high \(q_{\mathrm{bnd}}\) indicates that most of the mask lies in the image interior, whereas a lower value indicates that more mask pixels fall near the image boundary and the observation is more likely to be truncated. The three components are defined as
\begin{align}
q_{\mathrm{bnd},i}^{t}
&=
\frac{N_{\mathrm{int},i}^{t}+0.5}
{N_{\mathrm{tot},i}^{t}+1},
\nonumber\\
q_{\mathrm{pts},i,m}^{t}
&=
\frac{\min\!\left(N_{\mathrm{cand},i,m}^{t},N_{\mathrm{ref}}^{m}\right)+0.5}
{N_{\mathrm{ref}}^{m}+1},
\label{eq:q_components}\\
q_{\mathrm{clu},i,m}^{t}
&=
\frac{N_{\mathrm{clu},i,m}^{t}+0.5}
{N_{\mathrm{cand},i,m}^{t}+1}.
\nonumber
\end{align}
Here, $N_{\mathrm{tot},i}^{t}$ is the total number of pixels in the instance mask, and $N_{\mathrm{int},i}^{t}$ counts those outside an 8-pixel band along the image border. For modality $m$, $N_{\mathrm{cand},i,m}^{t}$ is the number of candidate 3D points before clustering, while $N_{\mathrm{clu},i,m}^{t}$ is the number of points in the cluster selected in Sec.~IV-C. The reference count $N_{\mathrm{ref}}^{m}$ is fixed at 16 for both modalities and shared across object categories, setting the saturation level of $q_{\mathrm{pts}}$. The additive constants smooth the ratios.

Each modality is screened independently using $q_{\min}=0.40$. If only one modality passes the threshold, its correction region is used directly; if both pass, their regions are merged by union. The larger accepted quality score is used as the observation weight:
\begin{equation}
R_i^t
=
\bigcup_{m:q_{i,m}^t\ge q_{\min}}
R_{i,m}^t,
\qquad
q_i^t
=
\max_{m:q_{i,m}^t\ge q_{\min}}
q_{i,m}^t.
\label{eq:modality_admission}
\end{equation}
The union retains complementary spatial support, while the maximum prevents a weaker modality from reducing the observation weight. If neither modality passes the threshold, no region is admitted to the temporal update.

\subsection{Cross Frame Patch Maintenance}
After quality filtering, each accepted object observation in the current frame provides a correction region \(R_i^t\) and its quality \(q_i^t\). Before a patch is created, the region is associated with a persistent object instance. The system first considers existing instances of the same object category and determines the association using region overlap, centroid distance, and minimum inter region distance. If no spatially consistent instance is found, a new instance ID is created. Regions associated with the same instance ID are then maintained together across frames. An observation rejected only by the quality threshold may still be associated with an existing instance, but it does not contribute to patch formation or refresh the patch lifetime.

For instance \(j\), the accepted regions associated with its ID are accumulated within a \(1.5\,\mathrm{s}\) window \(W_j^t\), with at least \(0.20\,\mathrm{s}\) between evidence samples. Their accumulated evidence is
\begin{equation}
E_j^t=\sum_{r\in W_j^t}q_r .
\end{equation}
The same regions are used to determine which grid cells should be retained. For each cell, the quality scores of the regions containing that cell are accumulated, and cells with sufficient support are kept. These cells together form the region to be committed. When \(E_j^t\ge E_{\min}=1.0\), the region is confirmed and written to the map through \textsc{COMMIT}, becoming the correction patch of that instance.

\begin{figure}[t]
  \centering
  \includegraphics[width=\columnwidth]{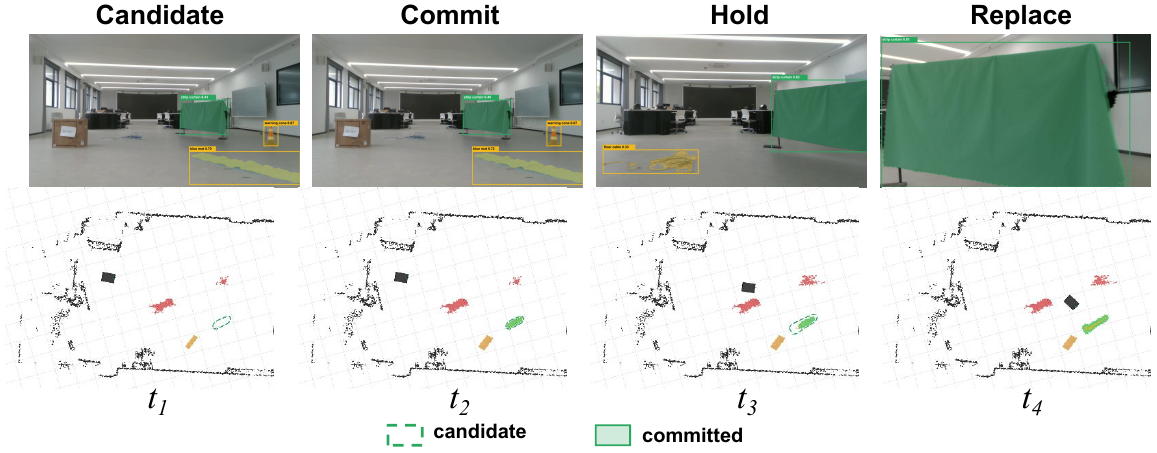}
  \caption{Cross frame maintenance of a REMOVE patch under changing viewpoints. A region accepted at \(t_1\) is retained as a candidate before sufficient evidence is accumulated, and is committed as a patch at \(t_2\). The patch is retained at \(t_3\) while a newly observed region is accumulated as a candidate, and is replaced at \(t_4\). Dashed outlines denote candidate regions under evidence accumulation, whereas filled regions denote committed patches.}
  \label{fig:memory}
\end{figure}

Once a patch has been committed, later regions associated with the same instance ID are compared with the current patch. If their difference remains within the change thresholds, the existing patch is retained. A region is considered substantially different when its centroid displacement is at least \(0.20\,\mathrm{m}\), its relative area change is at least \(50\%\), or its IoU with the patch is below \(0.35\). In this case, the existing patch remains unchanged while the new regions accumulate evidence. Once the new region is confirmed, it replaces the previous patch through \textsc{REPLACE}, while the instance ID remains unchanged. Each valid update refreshes the \(20\,\mathrm{s}\) lifetime; if no valid update is received before it expires, the patch is removed through \textsc{REVOKE}. Fig.~\ref{fig:memory} illustrates the \textsc{COMMIT}, hold, and \textsc{REPLACE} process under changing viewpoints.

\subsection{Costmap Integration}
\label{sec:map_fusion}
All active correction patches are overlaid on the navigation costmap. The corrected costmap is published at $5\,\mathrm{Hz}$ during navigation. \ADD creates obstacle regions and \EXTEND creates exclusion regions; both are mapped to non-traversable cells. \REMOVE clears spurious blockage from the online obstacle layer. When multiple patches affect the same cell, the non-traversable state produced by \ADD and \EXTEND takes precedence over \REMOVE. \REMOVE modifies only the online obstacle layer and does not override occupied or unknown cells in the static map.

\section{Experiments}

\subsection{Experimental Setup}
Real robot experiments are conducted on a Ranger Mini V3 four wheel steering mobile platform equipped with a Livox MID360S LiDAR and an Intel RealSense D435 RGB-D camera. Global paths are generated by a Hybrid A* planner~\cite{dolgov2010hybridastar}, and local trajectories are produced by TEB~\cite{rosmann2017teb}. All perception, map update, planning, and control modules except Qwen3-vl-flash run on an onboard computer equipped with an Intel Core i9-13900HX CPU, 32 GB RAM, and an NVIDIA GeForce RTX 4090 Laptop GPU (16 GB). The costmap uses a resolution of $0.05\,\mathrm{m}$ and an inflation radius of $0.05\,\mathrm{m}$. The maximum forward speed of the robot is limited to $0.8\,\mathrm{m/s}$.

\begin{figure*}[!t]
  \centering
  \includegraphics[width=\textwidth]{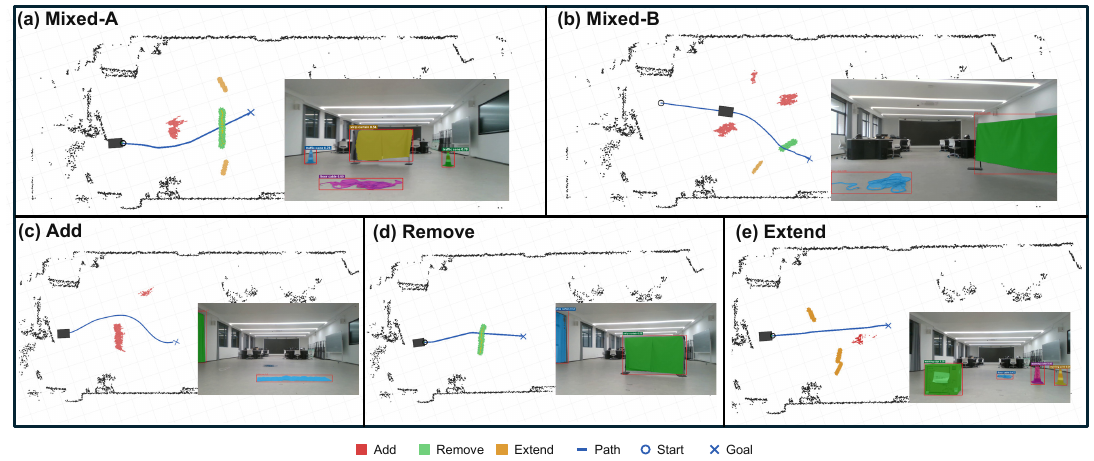}
  \caption{Real robot navigation in five layouts. Panels (a) and (b) show Mixed-A and Mixed-B, while (c), (d), and (e) focus on \ADD, \REMOVE, and \EXTEND, respectively. Each panel shows the corrected map and a representative navigation path, with an RGB inset displaying detected object instances.}
  \label{fig:real_eval}
\end{figure*}

The real world evaluation contains five fixed layouts, Mixed-A, Mixed-B, ADD-focused, REMOVE-focused, and EXTEND-focused, each with a fixed object arrangement and navigation goal. Mixed-A contains two traffic cones, one curtain, and one floor cable. Mixed-B contains an angled curtain, one foam board, one floor cable cluster, one traffic cone, and an additional box. The ADD-focused layout contains one foam board and two floor cables, the REMOVE-focused layout contains one curtain, and the EXTEND-focused layout contains two traffic cones and a box carrying a no entry sign, together with one floor cable. In the focused layouts, one correction action dominates the route, whereas Mixed-A and Mixed-B contain objects requiring different actions along the same route. We first evaluate the complete \NavPatch system in 50 closed loop trials across these layouts. We then conduct an ablation study of the patch update design using four configurations, Frame-wise, Quality-only, Temporal-only, and \NavPatch. Each configuration is evaluated in 10 trials per layout, giving 50 trials per configuration and 200 trials in total. The 50 \NavPatch trials in the ablation study are the same trials used for the full system evaluation. The five experimental layouts are shown in Fig.~\ref{fig:real_eval}.
 
Across the five layouts, we annotate 17 navigation relevant object instances: 8 ADD, 3 REMOVE, and 6 EXTEND instances. For each object instance, we record its position in the map frame and its expected correction action, and use the RGB field of view to determine the valid evaluation interval in each run.

Closed loop navigation is evaluated using Success Rate (SR), defined as the proportion of runs that reach the goal without collision. In REMOVE layouts, the robot must also pass through the curtain rather than bypass it. Correction accuracy is evaluated against manually annotated ground truth using action consistent, one to one spatial matching with a tolerance of $0.25\,\mathrm{m}$. False Commit Rate (FCR) is the proportion of committed patches that fail to match a ground truth object in action and position. Missed Commit Rate (MCR) is the proportion of visible ground truth objects for which no correct patch is committed during a run. Patch Update Rate (PUR) is the proportion of routed observations that change the committed patch. Overall rates are computed from pooled counts across runs.

\subsection{Real Robot Evaluation}
Table~\ref{tab:real_nav} summarizes the performance of the complete system across the five layouts. \NavPatch completes 43 of 50 trials with an overall SR of $86.0\%$. The focused layouts achieve $90.0\%$--$100.0\%$ SR. The maps in Fig.~\ref{fig:real_eval}(a)--(e) illustrate the different corrections used in these layouts.

The mixed layouts contain multiple objects requiring all three map actions within the same run. \NavPatch completes 7 of 10 trials in Mixed-A and 7 of 10 in Mixed-B. These layouts account for 6 of the 7 unsuccessful trials and have the highest MCR values, $35.0\%$ and $34.7\%$, respectively. Panels (a) and (b) of Fig.~\ref{fig:real_eval} show how the different corrections are combined in a shared costmap used for navigation.

Across the 50 trials, 42 object-level missed commits are recorded, including 13 \ADD, 10 \REMOVE, and 19 \EXTEND cases. In many cases, the VLM provides the relevant category and action, but GroundingDINO fails to localize the corresponding object reliably, preventing a matching correction patch from being committed. The camera view may also change before the next VLM output becomes available, while GroundingDINO continues using the previous detection phrases, which can produce responses in unrelated image regions. Many false commits are associated with distant objects, where the target occupies fewer image pixels and GroundingDINO becomes less reliable for instance localization. These errors increase FCR and MCR, but many occur outside the traversed corridor and therefore do not directly affect navigation.

\begin{table}[t]
\caption{Navigation and map correction results across five layouts}
\label{tab:real_nav}
\centering
\footnotesize
\setlength{\tabcolsep}{3.0pt}
\begin{tabular}{lrrrrr}
\toprule
Layout & Trials & SR & FCR  & MCR & PUR \\
\midrule
Mixed-A         & 10 & $70.0\%$ & $42.2\%$ & $35.0\%$ & $8.1\%$ \\
Mixed-B         & 10 & $70.0\%$ & $37.3\%$ & $34.7\%$ & $8.5\%$ \\
ADD-focused     & 10 & $100.0\%$ & $47.3\%$ & $3.3\%$  & $7.8\%$ \\
REMOVE-focused  & 10 & $100.0\%$ & $27.3\%$ & $20.0\%$ & $1.7\%$ \\
EXTEND-focused  & 10 & $90.0\%$  & $38.5\%$ & $20.0\%$ & $5.4\%$ \\
\midrule
Overall & 50
& $\mathbf{86.0\%}$
& $\mathbf{40.7\%}$
& $\mathbf{24.9\%}$
& $\mathbf{6.9\%}$ \\
\bottomrule
\end{tabular}
\end{table}

\subsection{Ablation Study of Patch Update Strategies}
To analyze the effects of observation quality filtering and cross frame maintenance of instance patches, we compare four configurations. \textbf{Frame-wise} directly updates the map using the correction region from the current instance observation. \textbf{Quality-only} adds observation quality filtering to frame-wise updates. \textbf{Temporal-only} uses cross frame patch maintenance without quality filtering. \textbf{NavPatch} combines observation quality filtering with cross frame patch maintenance, using quality scores as observation weights. All four configurations share the same visual semantic front end, 3D point extraction pipeline, and navigation back end.

\begin{table}[t]
\caption{Ablation study of patch update strategies}
\label{tab:patch_analysis}
\centering
\footnotesize
\setlength{\tabcolsep}{4.3pt}
\begin{tabular}{lcccc}
\toprule
Configuration & SR  & FCR & MCR  & PUR \\
\midrule
Frame-wise & $70.0\%$ & $68.4\%$ & $22.4\%$ & $95.4\%$ \\
Quality-only & $66.0\%$ & $64.6\%$ & $24.1\%$ & $88.6\%$ \\
Temporal-only & $54.0\%$ & $54.1\%$ & $27.6\%$ & $5.9\%$ \\
NavPatch & $\mathbf{86.0\%}$ & $\mathbf{40.7\%}$ & $24.9\%$ & $6.9\%$ \\
\bottomrule
\end{tabular}
\end{table}

Table~\ref{tab:patch_analysis} summarizes the ablation results. \NavPatch achieves the highest SR ($86.0\%$) and the lowest FCR ($40.7\%$), while maintaining a low PUR of $6.9\%$. Frame-wise obtains the lowest MCR ($22.4\%$), but its higher FCR and PUR indicate that correct patches are more susceptible to subsequent observation changes. Quality-only slightly improves FCR and PUR, whereas Temporal-only produces more stable updates but lower SR. Overall, the full configuration provides a better balance among object coverage, commit reliability, and patch stability.

These results reflect the different roles of the two mechanisms. Frame-wise directly updates the map using the current observation, making correct patches easier to obtain but also allowing single frame errors and fluctuations to affect the map. Quality-only rejects some observations with poor geometric support, but accepted observations are still updated frame by frame. Temporal-only suppresses transient updates through cross frame evidence accumulation and patch retention, but a more stable patch is not necessarily correct, and valid corrections may not be committed in time. \NavPatch further filters observations by quality and uses quality-weighted evidence across frames, allowing patch commitment and replacement to rely more on well supported observations and thereby improving both update stability and commit reliability.

\subsection{Runtime}
Runtime is evaluated across the 50 \NavPatch runs. For each metric, we first compute the median within each run and then report the median across runs. The remote VLM call takes $1.801\,\mathrm{s}$, while GroundingDINO and SAM~2.1 together take $131.1\,\mathrm{ms}$. The interval from receipt of a detection result to the associated semantic map commit message is $101.9\,\mathrm{ms}$. The remote VLM call is the slowest measured stage, whereas detection to commit updates have a median latency below $0.3\,\mathrm{s}$.

\section{Conclusion}
\NavPatch uses periodic VLM scene understanding to propose \ADD, \REMOVE, and \EXTEND corrections during navigation. Open vocabulary instance localization and LiDAR and RGB-D observations convert these proposals into local correction regions. Observation quality filtering and cross frame patch maintenance control patch commitment and replacement. In an ablation study of four configurations with 200 runs across five layouts, \NavPatch achieves an SR of $86.0\%$ and an FCR of $40.7\%$, compared with $70.0\%$ and $68.4\%$, respectively, for frame-wise updates. Remaining errors in instance localization can lead to missed or false commits.

\bibliographystyle{IEEEtran}
\bibliography{references}

\end{document}